\documentclass[%
 preprint,
 amsmath,amssymb,
]{revtex4-1}

\usepackage{graphicx}
\usepackage{amssymb}
\usepackage{setspace}
\usepackage[utf8]{inputenc}
\usepackage{array}
\usepackage{natbib}
\usepackage{amsmath}
\usepackage{multirow}
\usepackage{hyperref}% add hypertext capabilities
\usepackage{rotating}
\usepackage{adjustbox}
\usepackage{threeparttable}
\usepackage{booktabs}

\usepackage{color}
\definecolor{redcolor}{rgb}{1.0,0.,0.}

\definecolor{bluecolor}{rgb}{0,0.,1}

\begin{document}

\preprint{}

\title{Probing the statistical properties of enriched co-occurrence networks}% Force line breaks with \\

\author{Diego R. Amancio$^1$, Jeaneth Machicao$^2$ and Laura V. C. Quispe$^1$}

\affiliation{
$^1$Institute of Mathematics and Computer Science, University of S\~ao Paulo, S\~ao Carlos, Brazil\\\\
$^2$Escola Polit\'ecnica da Universidade de S\~ao Paulo (EPUSP), São Paulo, Brazil \\
}

\newpage 

\begin{abstract}
Recent studies have explored the addition of virtual edges to word co-occurrence networks using word embeddings to enhance graph representations, particularly for short texts. While these enriched networks have demonstrated some success, the impact of incorporating semantic edges into traditional co-occurrence networks remains uncertain. In this study, we investigate two key statistical properties of text-based network models. First, we assess whether network metrics can effectively distinguish between meaningless and meaningful texts. Second, we analyze whether these metrics are more sensitive to syntactic or semantic aspects of the text. Our results show that incorporating virtual edges can have both positive and negative effects, depending on the specific network metric. For instance, the informativeness of the average shortest path and closeness centrality improves in short texts, while the clustering coefficient's informativeness decreases as more virtual edges are added. Additionally, we found that including stopwords affects the statistical properties of enriched networks.
Our results can serve as a guideline for determining which network metrics are most appropriate for specific applications, depending on the typical text size and the nature of the problem.
\end{abstract}

\maketitle

%\linenumbers

% \doublespacing

%\input{sections/intro.tex}

\section{Introduction}

Word co-occurrence networks have been widely used in various text analysis studies, including authorship attribution~\citep{10.1371/journal.pone.0170527}, distinguishing real text from generated text~\citep{Amancio2015}, summarization~\citep{Tixier2016}, and language clustering~\citep{Liu2013, Vera_2021}. In these networks, words from a text are represented as nodes, with edges connecting adjacent words~\citep{Cancho2001}. However, this approach requires large textual corpora to provide sufficient graph-based data. Without enough data, the network topology becomes too simple, reducing its effectiveness in classification tasks.

The recent introduction of semantic edges (also known as virtual edges) enhances the discriminability of networks derived from shorter texts by avoiding linear structures~\citep{DBLP:journals/corr/SantosCOAMA17}. Virtual edges are hypothetical connections between non-adjacent nodes, based on the notion that semantically similar words can be linked. This enrichment of word co-occurrence networks, achieved by incorporating virtual edges, can be implemented using word embeddings, where similar words have similar embeddings.
This approach has shown significant advances in text classification~\citep{DBLP:journals/corr/SantosCOAMA17, QUISPE2021125344} and keyword extraction~\citep{GARG2018698, Tohalino2024}. However, the implications of integrating virtual edges into word co-occurrence networks remain still unexplored. 

We propose to analyze the effects of integrating virtual edges into word co-occurrence networks by examining two important properties of the metrics derived from these networks. The first property, referred to as informativeness, is the ability of the metrics to distinguish between real and meaningless texts. The second property aims to determine whether the metrics are more sensitive to syntactical/stylistic or semantic textual factors~\citep{amancio2013probing}. Although it is well-established that most metrics derived from traditional co-occurrence networks are informative and more dependent on syntax and style than on semantics, no study has investigated whether this property holds or varies with the number of edges added to the model. Motivated by this gap, this work is driven by the following research questions:

\begin{enumerate}   

\item How do virtual edges impact the informativeness of the metrics and their ability to differentiate between meaningful and nonsensical text?

\item How do virtual edges affect network features in capturing syntactic or semantic aspects of texts?

\item Do the answers to the above questions depend on the pre-processing steps in network construction, such as stopword filtering?

\end{enumerate}

Our analysis revealed that including virtual edges can indeed affect the statistical properties of networks, particularly in shorter texts. In these cases, the informativeness of metrics such as average shortest path length and closeness centrality is enhanced. Conversely, some metrics, like the clustering coefficient, experience a decrease in informativeness. Regarding the ability to capture syntactic features, our analysis showed that virtual edges in short texts increase the sensitivity of the average shortest path to semantics, while metrics like eigenvector centrality are largely unaffected and show no clear preference for syntactic or semantic features. In some cases, the sensitivity of metrics can shift from syntax to semantics with the addition of virtual edges, as observed with betweenness. We also found that filtering stopwords can affect the informativeness of metrics.

\section{Related Works}

Complex networks have been used to analyze texts in various contexts, including uncovering language patterns and performing text classification tasks~\citep{10.1371/journal.pone.0066344,10.1371/journal.pone.0214863,STANISZ20241,CONG2014598,WACHSLOPES20168,Cancho2001,PhysRevE.69.051915,Liu2010,GAO2014579,LIANG20094901,PhysRevE.69.051915,amancio2012using,amancio2016network}. 
The most commonly used network for text analysis is the word co-occurrence model. The work conducted in~\cite{GAO2014579} applied co-occurrence networks to analyze six languages, revealing language-specific patterns. The study compared the co-occurrence networks of Chinese and English texts, including essays, novels, articles, and reports. They found that all the networks exhibited scale-free and small-world properties. In particular, in networks derived from Chinese texts, the average shortest path length could distinguish between different content types, making it a style-dependent measure. Additionally, Chinese texts exhibited higher clustering coefficients, while English texts generally had shorter average path lengths.
The authors also found that Chinese networks are assortative, while English networks are disassortative.

In~\cite{PhysRevE.69.051915}, the authors analyzed patterns in syntactic dependency networks, where connections are based on syntactic dependencies. They demonstrated that networks derived from different languages—Czech, German, and Romanian—share complex statistical properties, such as the small-world phenomenon, degree distribution scaling, and disassortative mixing. Additionally, the authors noted that co-occurrence networks provide a simplified approach to constructing syntactic networks, as most syntactic links occur between adjacent words.

Co-occurrence words have also been used to study the statistical properties of unknown manuscripts. Using a co-occurrence strategy without any pre-processing step, the authors in~\cite{amancio2013probing} aimed to determine whether texts, such as the Voynich Manuscript, exhibit characteristics of natural language or are merely random character sequences. By analyzing word frequency, intermittency, and network properties, the authors provided valuable insights into the nature of the text, contributing to the understanding of unknown or undeciphered writings. Additionally, this research proposed a framework for evaluating the statistical properties of network models, including the ability of network metrics to distinguish between real and nonsensical texts. Furthermore, the study examined which linguistic properties -- syntactic or semantic --the metrics are likely to capture. This framework forms the foundation for the analysis conducted in this study.

More recently, enriched word co-occurrence networks have been proposed to capture information that traditional word co-occurrence networks may miss~\citep{CORREA2019180,DBLP:journals/corr/SantosCOAMA17,QUISPE2021125344,Tohalino2024}. \cite{DBLP:journals/corr/SantosCOAMA17} employed an enriched network to identify cognitive disorders from short text analysis by incorporating virtual edges. Including virtual edges was crucial not only for capturing additional linguistic information from the texts but also for creating a more complex topology in networks derived from short transcripts. 
Using a similar strategy, word embeddings and community detection were used for the problem of word sense induction, outperforming competing algorithms and baselines~\cite{CORREA2019180}. 

Enriched networks have also been applied to stylistic tasks, such as authorship recognition~\citep{QUISPE2021125344}. Upon employing different strategies to threshold the networks, the study evaluated various word embedding techniques, including Word2Vec, GloVe, and FastText~\citep{Mikolov2013,Pennington2014,bojanowski2017enriching}. The authors found that combining FastText with a global strategy yielded the best performance for short texts in the context of authorship recognition. In a similar approach, the study conducted in~\cite{Tohalino2024} analyzed the BERT model~\citep{devlin-etal-2019-bert} for adding virtual edges in networks for keyword extraction, achieving better results compared to other word embedding models.

While most studies involving enriched networks apply the model to various tasks, this study investigates how including virtual edges affects the statistical properties of the resulting networks. In addition to determining whether these networks can detect gibberish text, we evaluate, metric by metric, whether the model is more sensitive to stylistic/syntactic features or semantic aspects of the texts.

\section{Materials and Methods}

The methodology adopted in this work to analyze the statistical properties of enriched co-occurrence networks consists of two main phases: (i) network creation and (ii) topology characterization and analysis. In the network creation phase, we first pre-process the text, followed by constructing the network and adding virtual edges, effectively enriching the network with virtual links. During the topology characterization and analysis phase, metrics are extracted and classified based on two criteria: (i) \emph{informativeness}, referring to the metric's ability to distinguish between meaningful and meaningless texts, and (ii) \emph{variability ratio}, which evaluates whether the metric is more sensitive to syntactic or semantic features of the text. %The entire process is illustrated in Figure \ref{fig:virtualedges}. 

{\subsection{Network Construction}}

To construct networks from texts, we employed several pre-processing steps, including tokenization (i.e., identifying individual words), removing punctuation, and filtering out irrelevant information such as spaces and non-alphabetic characters.
The removal of stop-words is considered an option; therefore, we have two types of text: one with the complete text including stop-words and the other with stopwords removed. %The benefit of eliminating stop-words is also evaluated in the context of short texts, where retaining crucial %twork with filtering stop-words has fewer nodes than the other network.
{
After pre-processing, the texts are transformed into co-occurrence networks. In these networks, each unique word in the text becomes a node, and edges are created between nodes representing words that appear adjacent to each other in the text.}
The total number of edges in the network at the end of this phase is denoted by $N_E$.

%\textcolor{blue}{\subsubsection{Virtual Edge Addition}}

Originally, word co-occurrence networks were designed to connect only adjacent words. However, this approach proved inefficient for modeling shorter texts. Recent research has shown that incorporating virtual edges -- connections between similar but non-adjacent words -- provides valuable information, enabling more effective analysis of shorter texts~\citep{CORREA2019180}. In a co-occurrence network extracted from very short text, the topology is almost linear. The inclusion of virtual edges increases the complexity of the network, enhancing its usefulness for classification tasks~\citep{CORREA2019180}.

\begin{figure}[hbtp!]
    \centering
    \includegraphics[width=1.0\linewidth]{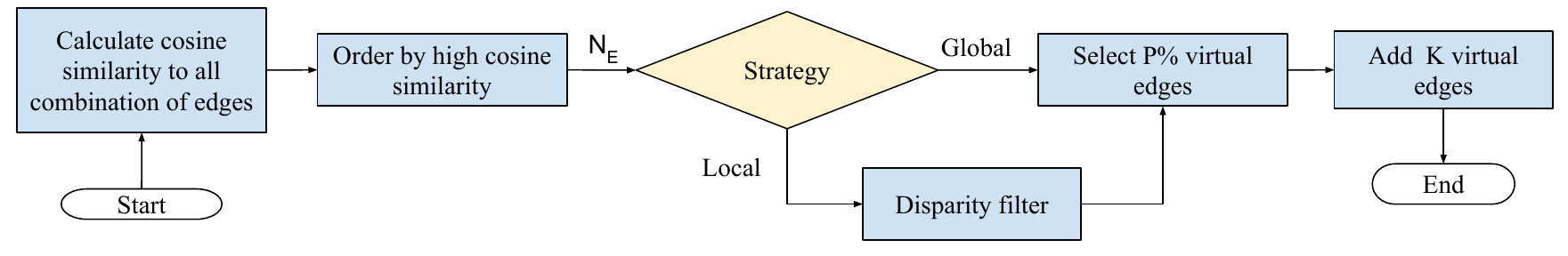}
    %\captionsetup{format=plain}
    \caption{Methodology used in this study analyzes enriched networks. All pairs of similarities are calculated, and the similarity weights are sorted in decreasing order. To filter the edges, the global strategy selects those with the highest weights across the entire network, while the local strategy evaluates the importance of an edge based on the local structure of each node. The total number of included edges is a parameter that varies throughout the analysis.}
    \label{fig:virtualedges}
\end{figure}

Figure~\ref{fig:virtualedges} illustrates the process of adding virtual edges to enrich the network's structure with semantic information. 
First, we obtain the word embeddings for each word in the network and calculate the cosine similarity between words that are unconnected in the original co-occurrence model (i.e., the co-occurrence network without virtual edges). This cosine similarity then becomes the weight of the potential virtual edge.
While all these pairs of nodes represent possible virtual edges that could be added to the network, including all of them would result in a network (clique) that lacks meaningful topological information, especially when analyzing unweighted metrics. To address this, we applied different criteria to retain only the most significant links: the global and local strategies.

\begin{itemize}

    \item \emph{Global Strategy} $(\mathcal{G})$: potential virtual edges are ranked by their weights, and the top $K$ edges are selected for inclusion. According to~\cite{QUISPE2021125344}, $K$ is selected as a percentage $P$ of the total edges in the original co-occurrence network ($N_E$), i.e. $K = P N_E$. In this case, we selected the following values: $P=\{25\%, 50\%, 75\%, 100\%\}$. %Figure~\ref{fig:distribuition-
    
    \item \emph{Local Strategy} $(\mathcal{L})$:
    Unlike the global strategy, this method considers the local topology of the final network after adding virtual edges. It retains only the most significant virtual edges, removing non-relevant ones while preserving the key features of the network. This approach, known as the disparity filter~\citep{serrano2009extracting}, establishes a null model to quantify the probability of a node being connected to an edge with a given weight, considering its other connections. This probability is then used to evaluate the significance of the edge. Specifically, the significance of an edge $e_{ij}$ is measured by $\alpha_{ij}$:
    \begin{equation} \label{eq:alpha}
        \alpha_{ij} = 1 - (k_i-1)\int_{0}^{\pi_{ij}}(1-x)^{k_i-2} dx
    \end{equation}
    \begin{equation} \label{eq:prob}
        \pi_{ij} = {w_{ij} \Bigg{(} \sum_{{ik}\,\in\,E}w_{ik}} \Bigg{)}^{-1},
    \end{equation}
    where $w_{ij}$ is the weight of $e_{ij}$, $k_i$ is the degree of the $i$-th node and $E$ is the set of all edges connected to node $i$. The high significance of an edge is represented by low values of $\alpha_{ij}$.
    In this context, the weights of the edges are computed as the embedding similarity using cosine similarity (for virtual edges). To preserve the edges obtained from co-occurrence, we assigned them the maximum similarity value (i.e., 1).
    The disparity filter removes the edges with the lowest significance, leaving $K$ edges. For comparison, $K$ is set to match the number of edges included in the global strategy (i.e., $K$ is a fraction of the total edges in the co-occurrence network, excluding virtual edges).

    \end{itemize}

    We used the pre-trained FastText model~\citep{grave2018learning} to map words to embeddings. This model operates at the character level, removing the need for lemmatization. FastText is available for multiple languages and captures semantic information by projecting words into high-dimensional vector spaces (word embeddings) with 300 features as the vector size. We opted not to test other embedding models due to FastText's versatility across languages and its comparable performance to models like Word2Vec in practical applications involving enriched networks~\citep{QUISPE2021125344}.

\subsection{Network Analysis}

Once the networks with virtual edges are constructed and the top $K$ virtual edges are identified, the network can be analyzed using network metrics. The network analysis consists of the following steps:

\begin{enumerate}

    \item \emph{Network metrics extraction}: network metrics are extracted from the enriched networks.
    
    \item \emph{Metrics normalization}: to avoid bias in the metrics caused by network size, the extracted metrics are normalized.

    \item \emph{Statistical properties analysis}: this is the most important part of the study, as it assesses the informativeness of the metrics and determines whether they are more effective in capturing  syntactic or semantic features of texts.

\end{enumerate}

\subsubsection{Network metrics}

In our analysis, we focused on the most commonly used network metrics for analyzing networks derived from text. These metrics include average shortest path length ($L$), clustering coefficient ($C$), closeness centrality ($CC$), betweenness centrality ($B$), PageRank ($PR$), and eigenvector centrality ($EV$)~\citep{amancio2013probing}.
For metrics calculated at the node level, summarization is necessary. In other words, for each local metric computed for individual nodes in the network, we aim to condense the information into a single value that represents the entire network.
Following~\cite{amancio2013probing}, two types of summarization can be considered:
\begin{enumerate}
    \item The average measure across all nodes, denoted as $\tilde{X}$.
    \item The average measure for the most important words, denoted as $\tilde{X}^{*}$. The most important words are defined as the top 10 most frequent words in each text.
\end{enumerate}

\subsubsection{Metrics normalization}

Network metrics derived from texts can have their metrics dependent on text size (such as the number of tokens or vocabulary size). 
For this reason, we adopted the following procedure to normalize the metrics. For each original text, we generated 10 shuffled versions, where the shuffling process is performed at the word level. Let $\tilde{X}^{R}$ represent the value of a specific measure computed for each shuffled text. If $\mu{(\tilde{X}^{R})}$ denotes the average value computed across the shuffled versions, we define the normalized value of the metric, $X$, as
\begin{equation}
    X= \frac{\tilde{X}}{\mu{(\tilde{X}^{R}})}.
\end{equation}
Considering the inherent uncertainty in the random shuffling process, the error associated with $X$ is quantified as
\begin{equation}
    \epsilon(X) = \frac{\sigma(\tilde{X}^{R})}{\mu{(\tilde{X}^{R})}}X,
\end{equation}
where $\sigma(\tilde{X}^{R})$ is the standard deviation computed over the shuffled versions.

\subsubsection{Statistical properties analysis}

The first key property addressed in this paper is the \emph{informativeness} of the metric. A metric is considered informative if it can effectively differentiate between real and shuffled texts. This property is crucial as it indicates the metric's ability to identify texts where semantic words are combined in a nonsensical manner~\citep{amancio2013probing}. An informative metric is valuable not only for detecting such anomalies but also for capturing subtle nuances in text styles. If a metric is not informative, its normalized value is expected to be close to $X=1$. To quantify informativeness, we measure the distance ($D$) of $X$ from 1 for each text (network) and normalize this distance by the error $\epsilon(X)$, i.e.:
\begin{equation}
    D = \frac{X-1}{\epsilon(X)}.
    \label{eq:D}
\end{equation}
If the distance $D>1$, it indicates that the measure is informative for the text being analyzed.
To assess the informativeness of the dataset, we used a criterion that measures the proportion of texts where $D > 1$. This is defined as:
\begin{equation}
    I = \frac{|D > 1|}{N_T},
    \label{eq:I}
\end{equation}
where $|D > 1|$ denotes the number of texts for which the condition $D > 1$ is satisfied, and $N_T$ represents the total number of texts analyzed in the dataset.

In the context of enriched networks, analyzing informativeness is crucial because the inclusion of many virtual edges can diminish the prominence of the original co-occurrence edges, which store word ordering information. As a result, the network structure might only reflect a similarity network and lose informativeness completely. 

Another key aspect of the paper is its ability to capture \emph{syntactical} or \emph{semantic} aspects of language. The significance of this capability lies in determining the types of tasks where the metric could be applied. For instance, in tasks where style or language plays a crucial role, metrics that depend on syntax are more suitable. This is particularly relevant when identifying the nature of an unknown sequence of symbols or when determining the authorship of a text. Conversely, metrics that are more dependent on semantics could be used to identify, for example, shifts in semantic flow within texts, as in tasks like topic segmentation~\citep{CORREA2019180}.

To determine whether a measure $X$ is more dependent on linguistic structure (syntax) than on content (semantics), we used two datasets. The first dataset consists of the same text translated into different languages. For this, we used the New Testament, translated into different languages (NLANG dataset, as presented in Section \ref{sec:dataset}). This dataset allows us to assess the variability of a metric when the semantics remain constant, but the syntax changes with each language.
The second dataset was used to measure the variability of semantics across different texts in a single language. For this purpose, we used a collection of texts (novels) in English (NEN dataset, as presented in Section \ref{sec:dataset}).

By analyzing both datasets, we were able to measure, for each metric $X$, the variability across syntax and semantics. The variability across syntax is computed using the coefficient of variation $v_{(t=\textrm{nt},l)}$, where $t=\textrm{nt}$ indicates that the variation is computed for a specific text (the New Testament), and $l$ indicates that the dataset comprises texts in different languages (NLANG dataset). The variability across semantics is computed using the coefficient of variation $v_{(t,l=\textrm{en})}$. This means that we measured the variability within a dataset where the language is constant (English) and the textual content (i.e. the semantics) differ. This corresponds to the NEN dataset. 
To compute the nature of a metric $X$ we compute the following variability ration between $v_{(t=\tau,l)}$ and $v_{(t,l=\lambda)}$:
\begin{equation}
     V_R = \frac{v_{(t=\textrm{nt},l)}}{v_{(t,l=\textrm{en})}}.
\end{equation}

\begin{figure}[hbtp!]
    \centering
    \includegraphics[width=0.7\linewidth]{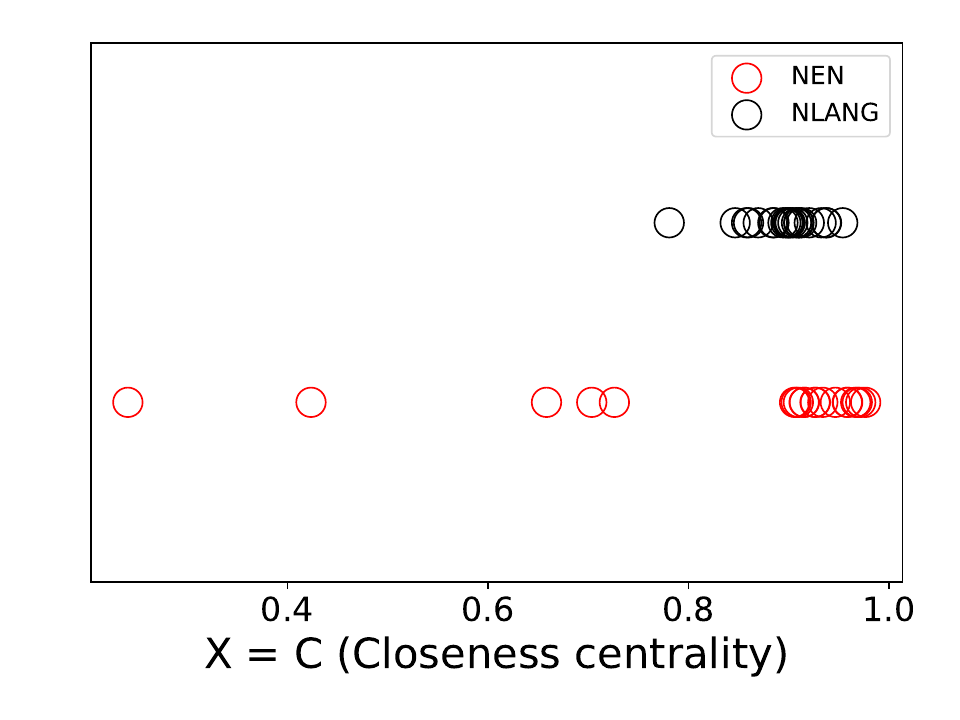}
    \caption{Example of hypothetical closeness centrality values obtained from the NEN and NLANG datasets. Because the variability in the NEN dataset is greater than that in the NLANG dataset, we have $V_R < 1$. This indicates that the metric is more dependent on semantics than on syntax.
}
    \label{fig:exa}
\end{figure}

$V_R$ can be used to determine whether a metric is more dependent on syntax or semantics. If $V_R > 1$, it indicates that the variability across syntax is greater than the variability across semantics, meaning the metric is more dependent on syntax~\citep{amancio2013probing}.
To illustrate this concept, Figure \ref{fig:exa} shows the behavior of the metric $X=C$ using hypothetical values from both the NEN and NLANG datasets. The figure clearly shows that the variability across texts in the same language   is greater than the variability of the same text across different languages. This suggests that the metric is more influenced by semantics than by syntax.

While most metrics in the traditional word co-occurrence networks are designed to capture syntactic relationships~\citep{amancio2013probing}, the addition of virtual (semantic) edges can alter the nature of these metrics. Therefore, it is important to analyze the properties of metrics extracted from enriched networks to determine if the metrics can effectively capture the intended features.

\subsection{Dataset} \label{sec:dataset}

%\textcolor{red}{lembrar de trocar por NLANG and NEN}
Two different datasets were used in this study, selected to enable direct comparison with previous research evaluating the statistical properties of texts modeled as traditional co-occurrence networks~\citep{amancio2013probing}.
The first dataset, referred to as the NEN dataset, was extracted from Project Gutenberg~\footnote{\url{http://www.gutenberg.org}} and consists of English subtexts extracted from various novels, with varying text lengths. The books included are David Copperfield (Charles Dickens); Dracula (Bram Stoker); Evelina, Or, the History of a Young Lady's Entrance into the World (Fanny Burney); Great Expectations (Charles Dickens); History of Tom Jones, a Foundling (Henry Fielding); Moby Dick; Or, The Whale (Herman Melville); Persuasion (Jane Austen); Pride and Prejudice (Jane Austen); The Life and Adventures of Robinson Crusoe (Daniel Defoe); and Ulysses(James Joyce). The text sizes used for each book are 200, 400, 800, and 1,000 words, corresponding to the first tokens of the texts. This dataset is used to measure the variability of metrics extracted from different texts in the same language.  

The second dataset, referred to as the NLANG dataset, consists of subtexts of varying sizes extracted from the New Testament of the Bible. The texts are translated into Arabic, English, Esperanto, German, Hebrew, Hungarian, Korean, Latin, Portuguese, Russian, and Vietnamese. This dataset is used to measure the variability of metrics extracted from the same content across distinct languages.

\section{Results and discussion}

%In this section, we evaluate the statistical properties of metrics extracted from enriched networks. We begin in Section \ref{secr1} by presenting the results for networks constructed without stopwords. The impact of including stopwords is then analyzed in Section \ref{secr2}.

\subsection{Informativeness and Variability Ratio Analysis} \label{secr1}

In this section, we analyze the behavior of informativeness and the variability ratio for the selected complex network metrics. Our analysis focuses on networks derived from texts \emph{without stopwords}. Additionally, we focus our discussion on the global thresholding strategy, as the local approach yielded similar results.% similar to the global perspective. 

\begin{figure}[hbtp!]
    \centering
    \includegraphics[width=0.95\linewidth]{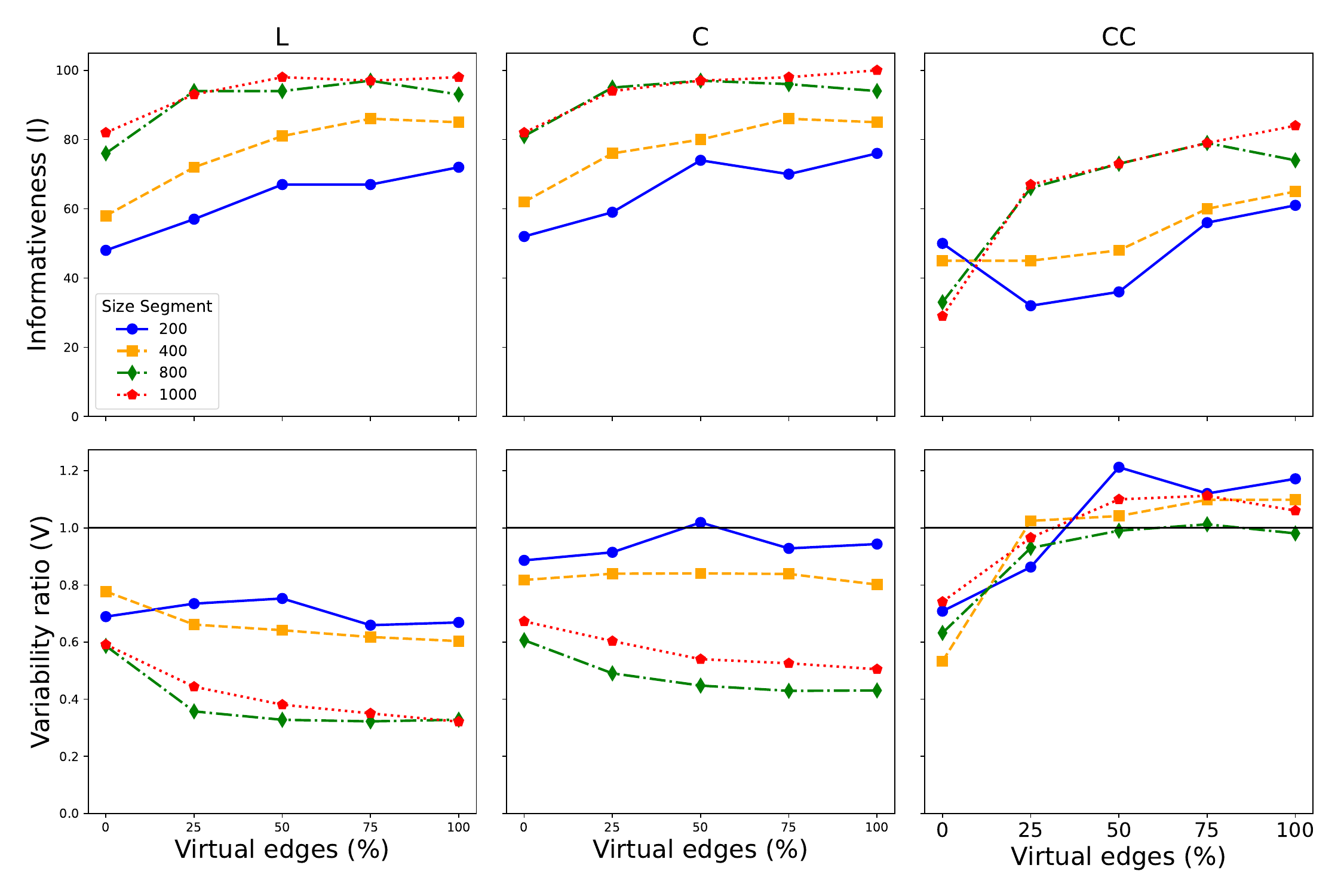}
    \includegraphics[width=0.95\linewidth]{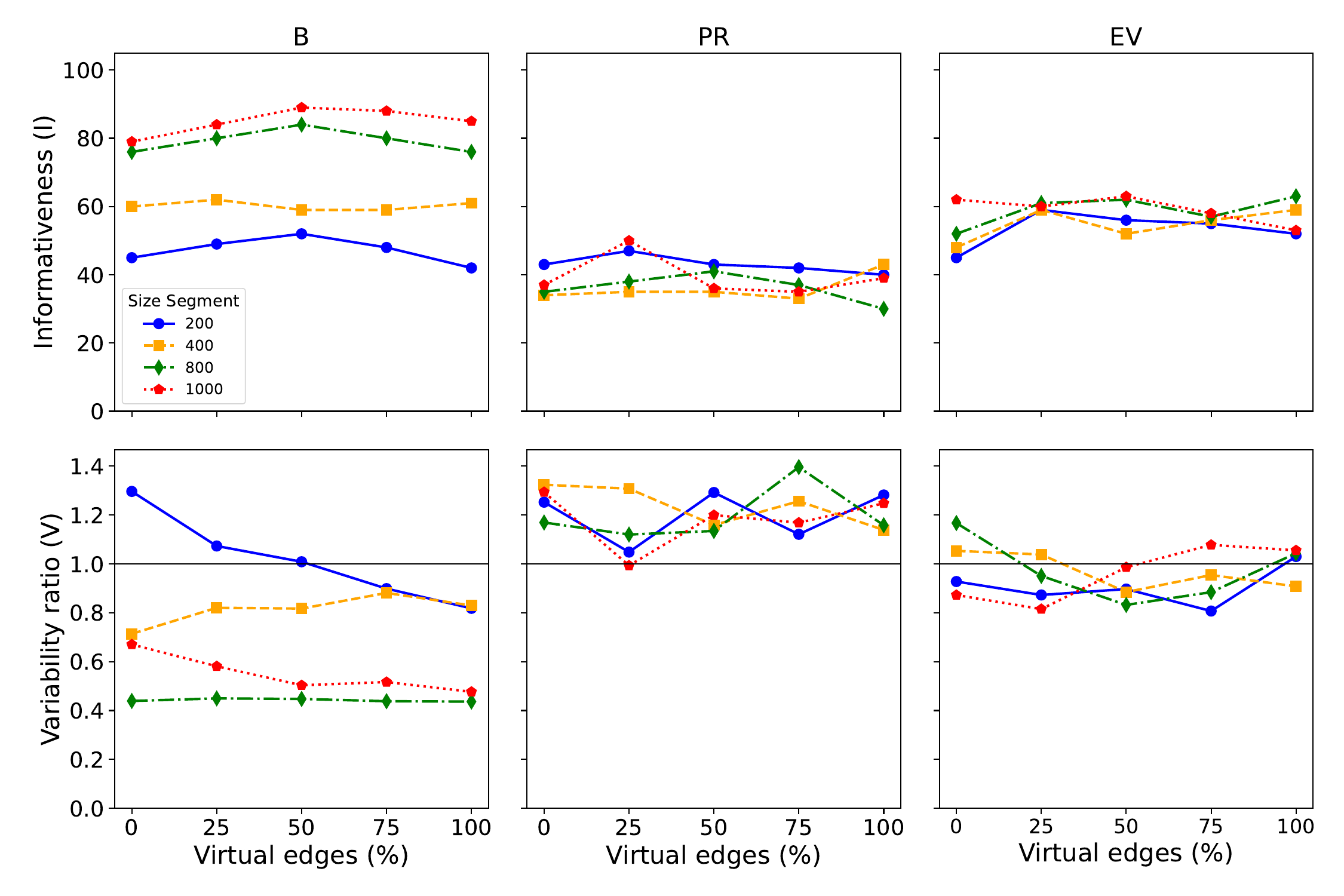}
    \caption{Global Strategy: Distribution of Informativeness and Variability measures for the Average Shortest Path (L), Closeness Centrality (C), Clustering Coefficient (CC), Betweenness Centrality (B), PageRank (PR), and Eigenvector Centrality (EV), with the addition of virtual edges in networks generated with variate text sizes and with filtering stop-word.}
    \label{fig:infovar-global_strategy}
\end{figure}

Figure~\ref{fig:infovar-global_strategy} illustrates the informativeness and variability ratio of network metrics, considering different text sizes and the network weight thresholding based on the global strategy. We begin by analyzing the network metrics computed across all the nodes in the network (i.e., $X$ as opposed to $X^*$).
The results obtained for the local strategy are shown {in Figure \ref{fig:infovar-local_strategy} of the Supplementary Information.} 

The main findings in Figure \ref{fig:infovar-global_strategy} are summarized below:

\begin{itemize}

    \item \emph{Average shortest path ($L$)}: The informativeness of $L$ clearly depends on text size. Without virtual edges, $L$ tends to be more informative for longer text segments. Adding virtual edges enhances the informativeness of the metric, especially for shorter texts. For texts longer than 800 tokens, the inclusion of virtual edges ensures that informativeness is maintained across all texts in the dataset. The variability ratio suggests that the normalized metric, as defined in our methodology, is more influenced by semantics than syntax in shorter texts. This effect becomes even more pronounced when virtual edges are included in shorter texts.

    \item \emph{Closeness centrality ($C$)}: The informativeness behavior is similar to the average shortest path length, as the metrics are correlated. However, regarding the variability ratio, in very short segments (200 tokens), there is no clear dominance in the ability to capture either syntax or semantics, even with the inclusion of virtual edges.

    \item \emph{Clustering Coefficient (CC)}: the clustering coefficient behaves differently from $L$ and $C$. For short texts, an increase in the number of edges can lead to a decrease in informativeness. However, for longer segments (over 400 tokens), the inclusion of edges enhances the informativeness metric. In these cases, $I$ can rise from approximately 30\% to 80\% when virtual edges are included. The variability ratio also shows interesting behavior. In very short texts, $CC$ seems more influenced by semantic features in traditional co-occurrence documents. However, with the addition of virtual edges, this behavior reverses, and $CC$ becomes more dependent on syntax.

    \item \emph{Betweenness Centrality (B)}: the inclusion of virtual edges has a minimal effect on informativeness. However, the results indicate that the betweenness informativeness for longer texts tends to be significantly higher than for shorter documents. Conversely, the behavior of the variability ratio depends on the size of the text. Larger texts remain unaffected, while for shorter texts, the variability ratio shifts from over 1.30 to 0.80 when 100\% of virtual edges are included, suggesting that the addition of edges increases the metric's sensitivity to semantics.

    \item \emph{PageRank (PR)}: the informativeness of PageRank consistently remains low across all text sizes, highlighting its limited sensitivity to incorporating virtual edges. This suggests that this metric may not be effective in distinguishing between gibberish and more detailed stylistic information in unweighted co-occurrence networks
    The variability ratio exhibits oscillatory changes with the addition of virtual edges showing a consistent dependence on syntax across all text sizes.

    \item \emph{Eigenvector Centrality (EV)}: the informativeness of this metric is similar to that of PageRank, though the values for $EV$ are slightly higher. Once again, the informativeness does not change significantly with the inclusion of virtual edges. Similarly, the variability ratio is also minimally affected by the addition of virtual edges. The variability ratio values suggest that there is no dominant dependence on either syntactic or semantic features.

\end{itemize}

We now analyze the network metrics that are computed exclusively based on the most frequent words in the text (i.e., $X^*$ as opposed to $X$). The results for the global thresholding strategy are presented in Figure \ref{fig:infovar-local_strategy}. The main results are summarized below:

\begin{figure}[hbtp!]
    \centering
    \includegraphics[width=0.95\linewidth]{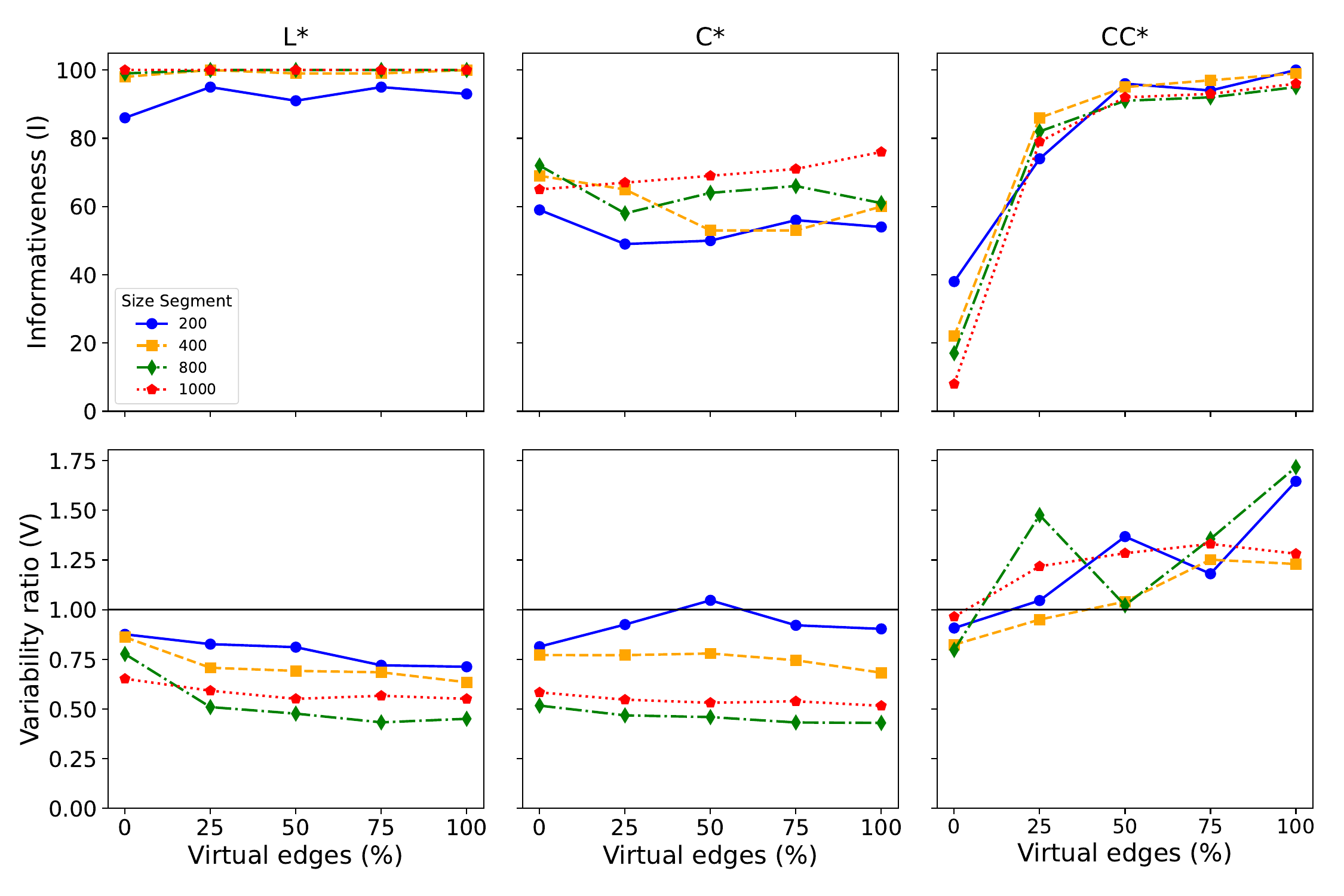}
    \includegraphics[width=0.95\linewidth]{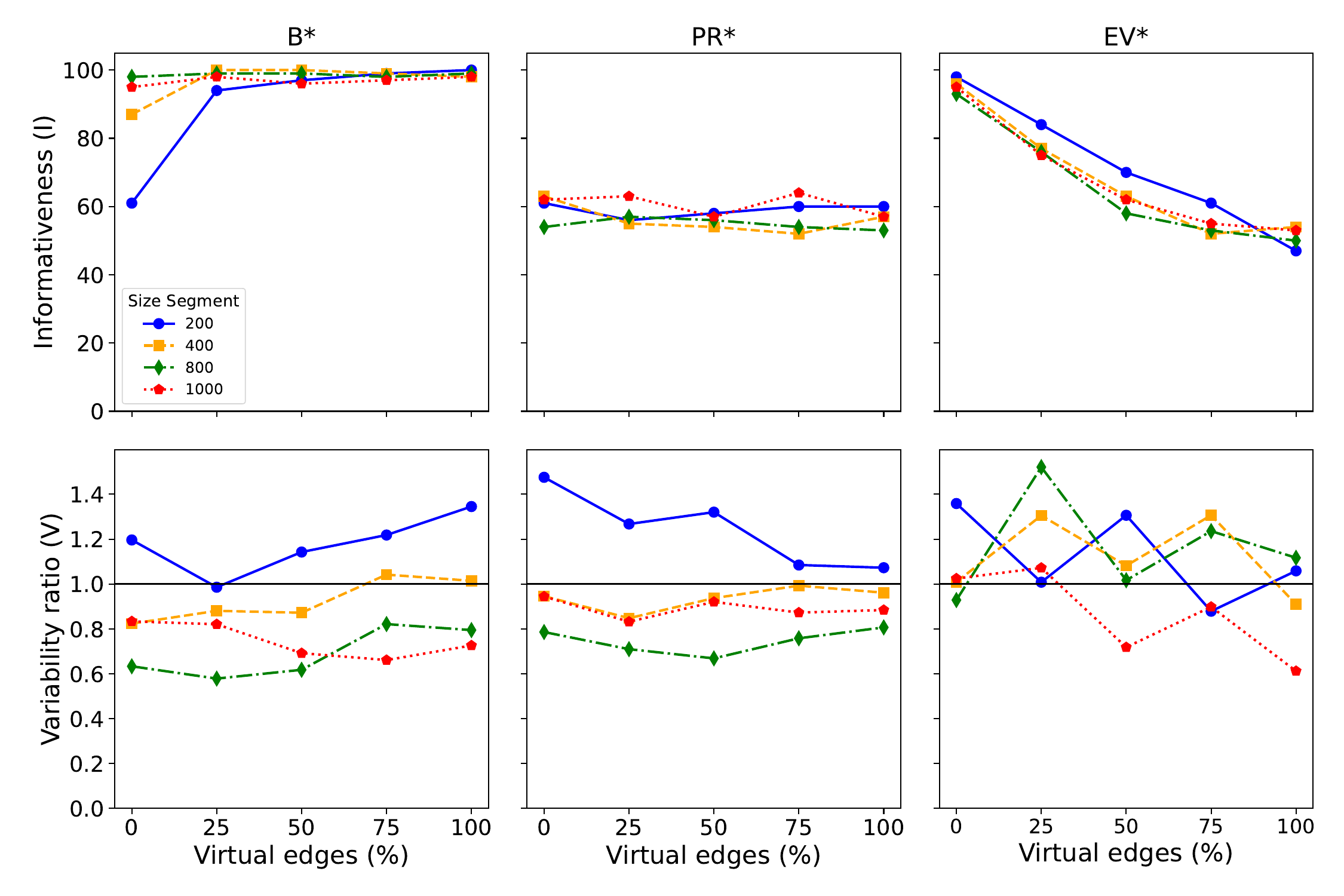}
    \caption{Global Strategy: Distribution of Informativeness and Variability measures for the Average Shortest Path (L*), Closeness Centrality (C*), Clustering Coefficient (CC*), Betweenness Centrality (B*), PageRank (PR*), and Eigenvector Centrality (EV*), with the addition of virtual edges in networks generated with variate text sizes and with filtering stop-word.}
    \label{fig:infovar-local_strategy}
\end{figure}

\begin{enumerate}
    
    \item \emph{Average shortest path ($L^*$)}: the informativeness of \(L^*\) is less sensitive to the inclusion of virtual edges compared to \(L\). However, very short text segments tend to be more affected. Despite this, \(L\) still maintains high informativeness even without virtual edges. The variability ratio also indicates that this feature is only weakly influenced by including virtual edges. Furthermore, the behavior of the curves for \(L^*\) closely resembles that of \(L\).
    This result also suggests that \(L^*\) is more influenced by semantics than syntax in shorter texts.

    \item \emph{Closeness centrality ($C^*$)}: the closeness shows little dependency with the inclusion of virtual edges, even for short texts. The informativeness for larger texts increases by a small margin. The variability ratio seems to be more dependent on text size than on the inclusion of virtual edges. Larger pieces of texts tend to capture more semantical features, similarly to the behavior of $C$. 

    \item \emph{Clustering Coefficient ($CC^*$)}: The inclusion of virtual edges has a significant impact on informativeness. Even a minimal inclusion can lead to a substantial increase in informativeness across all text sizes. Additionally, the variability ratio is influenced by the presence of virtual edges, with values exceeding the threshold of 1 for all text sizes. When compared to $CC$, the gain in informativeness appears to be stronger in this case.

    \item \emph{Betwenness ($B^*$)}:  The informativeness of betweenness primarily affects shorter texts. After adding 50\% virtual edges, all text sizes become 100\% informative. This behavior differs from what was observed with $B$, where the informativeness curve appeared to be strongly dependent on text size.
    Regarding the variability ratio, there is a clear dependence on text size. Differently from $B$, adding 50\% or more virtual edges to very short texts seems to increase the variability ratio.

    \item \emph{PageRank ($PR^*$)}: The informativeness of PageRank appears to remain unaffected by the inclusion of virtual edges, regardless of text size. This behavior is consistent with what has been observed for PR. Conversely, the variability ratio is more impacted in very short texts, leading to decreased variability.

    \item \emph{Eigenvector Centrality ($EV^*$)}: Unlike EV, informativeness is significantly affected. We observe that the informativeness values decrease sharply with the inclusion of virtual edges, and this trend is consistent across all text sizes. This contrasts with $EV$, where informativeness is only mildly affected. However, the variability ratio exhibits an oscillatory pattern, indicating this metric may be very sensitive to including virtual edges.

\end{enumerate}

\subsection{Impact of stop-words} \label{secr2}

In Table \ref{tab:comparision-strategy-part1}, We present the impact of including or filtering stopwords in the network analysis. Specifically, we focus on how the highest informativeness values—across varying numbers of virtual edges—are affected by the inclusion or exclusion of stopwords. 
Similarly, when analyzing the variability ratio, we analyze how the metric changes in the scenario where the network shows the greatest dependence on either syntax or semantics, corresponding to the minimum or maximum variability ratio values.
The analysis covers both global ($\mathcal{G}$) and local ($\mathcal{L}$) thresholding strategies. For comparison, we also provide the results for the original ($\mathcal{O}$) network (i.e., the network without virtual links). Additionally, the results are segmented by text size. We focus our discussion, however, on the shortest and longest text sizes.

\begin{table}[ht]
    \centering
    \caption{Comparison of informativeness and variability metrics among the Original network (O), Global strategy (G), and Local strategy (L), with and without stop-words filtering (SW). High values of informativeness measures in networks incorporating virtual edges are highlighted, along with their respective variability.}
    \label{tab:comparision-strategy-part1}
    %\scriptsize
    \begin{tabular}{lcc|cc|cc||cc|cc|cc}
        \hline
        &\multicolumn{6}{c}{Informativeness}&\multicolumn{6}{c}{Variability}\\
        \cline{2-13}
        &$\mathcal{O}$&$\mathcal{O}_F$&$\mathcal{G}$&$\mathcal{G}_F$&$\mathcal{L}$&$\mathcal{L}_F$&$\mathcal{O}$&$\mathcal{O}_F$&$\mathcal{G}$&$\mathcal{G}_F$&$\mathcal{L}$&$\mathcal{L}_F$\\
        \cline{2-13}
        &\multicolumn{12}{c}{Text size 200}\\
        \cline{2-13}
        L & 42.0 & 48.0 & 52.0 & 72.0 & 52.0 & 72.0& 2.3& 0.7& 1.4& 0.7& 1.4& 0.7\\
        C & 48.0 & 52.0 & 55.0 & 76.0 & 53.0 & 76.0& 2.3& 0.9& 1.5& 0.9& 1.5& 0.9\\
        CC & 47.0 & 50.0 & 76.0 & 61.0 & 76.0 & 61.0& 1.7& 0.7& 1.9& 1.2& 1.9& 1.2\\
        B & 40.0 & 45.0 & 51.0 & 52.0 & 51.0 & 54.0& 2.2& 1.3& 1.7& 1.0& 1.7& 1.1\\
        PR & 58.0 & 43.0 & 60.0 & 47.0 & 47.0 & 51.0& 1.1& 1.3& 1.1& 1.0& 1.0& 1.2\\
        EV & 46.0 & 45.0 & 66.0 & 59.0 & 67.0 & 52.0& 1.1& 0.9& 0.8& 0.9& 1.3& 1.0\\
        \hline
        &\multicolumn{12}{c}{Text size 400}\\
        \cline{2-13}
        L & 46.0 & 58.0 & 59.0 & 86.0 & 59.0 & 85.0& 2.1& 0.8& 1.6& 0.6& 1.6& 0.6\\
        C & 45.0 & 62.0 & 54.0 & 86.0 & 58.0 & 85.0& 2.1& 0.8& 1.6& 0.8& 1.7& 0.8\\
        CC & 46.0 & 45.0 & 83.0 & 65.0 & 83.0 & 65.0& 2.0& 0.5& 1.9& 1.1& 1.9& 1.1\\
        B & 43.0 & 60.0 & 53.0 & 62.0 & 53.0 & 61.0& 1.9& 0.7& 1.7& 0.8& 1.7& 0.8\\
        PR & 41.0 & 34.0 & 43.0 & 43.0 & 45.0 & 46.0& 0.9& 1.3& 1.1& 1.1& 0.9& 1.0\\
        EV & 54.0 & 48.0 & 74.0 & 59.0 & 76.0 & 59.0& 1.2& 1.1& 0.7& 1.0& 0.7& 0.9\\
        \hline
        &\multicolumn{12}{c}{Text size 800}\\
        \cline{2-13}
        L & 52.0 & 76.0 & 59.0 & 97.0 & 59.0 & 93.0& 1.9& 0.6& 1.2& 0.3& 1.2& 0.3\\
        C & 50.0 & 81.0 & 60.0 & 97.0 & 60.0 & 94.0& 2.0& 0.6& 1.3& 0.4& 1.3& 0.4\\
        CC & 38.0 & 33.0 & 96.0 & 79.0 & 96.0 & 74.0& 2.2& 0.6& 2.0& 1.0& 2.0& 1.0\\
        B & 45.0 & 76.0 & 60.0 & 84.0 & 58.0 & 77.0& 1.4& 0.4& 1.3& 0.4& 1.2& 0.4\\
        PR & 55.0 & 35.0 & 55.0 & 41.0 & 37.0 & 45.0& 1.2& 1.2& 1.2& 1.1& 1.3& 1.1\\
        EV & 56.0 & 52.0 & 78.0 & 63.0 & 83.0 & 63.0& 1.2& 1.2& 0.7& 1.0& 0.7& 1.0\\
        \hline
        &\multicolumn{12}{c}{Text size 1000}\\
        \cline{2-13}
        L & 68.0 & 82.0 & 74.0 & 98.0 & 74.0 & 99.0& 2.3& 0.6& 1.6& 0.4& 1.7& 0.4\\
        C & 62.0 & 82.0 & 73.0 & 100.0 & 72.0 & 100.0& 2.4& 0.7& 1.7& 0.5& 1.6& 0.5\\
        CC & 54.0 & 29.0 & 91.0 & 84.0 & 90.0 & 84.0& 2.0& 0.7& 1.7& 1.1& 1.5& 1.1\\
        B & 59.0 & 79.0 & 70.0 & 89.0 & 66.0 & 88.0& 1.8& 0.7& 1.6& 0.5& 1.6& 0.5\\
        PR & 41.0 & 37.0 & 41.0 & 50.0 & 41.0 & 46.0& 1.0& 1.3& 1.0& 1.0& 1.0& 1.0\\
        EV & 67.0 & 62.0 & 79.0 & 63.0 & 85.0 & 62.0& 1.5& 0.9& 0.8& 1.0& 0.8& 0.9\\
        \hline
    \end{tabular}
\end{table}

For texts of size 200 tokens, we observe that in the global strategy, filtering stopwords significantly enhances the informativeness of the $L$ and $C$ metrics, increasing their values by over 20\%. However, a decrease in informativeness is observed for both $PR$ and $EV$. The only metric that remains practically unaffected is betweenness, indicating that it is relatively stable regardless of the presence of stopwords. Using the local strategy does not seem to strongly affect the margin of gain (or loss) in terms of informativeness. In the global thresholding analysis, the largest changes in the variability ratio are observed for $L$, $CC$, and $B$. These metrics decrease in value and eventually shift their dependency from syntax to semantics.

For texts comprising 1000 tokens and using the global thresholding strategy, $L$, $C$, and $B$ show a significant increase in informativeness, while $PR$ experiences only a slight improvement. Conversely, $EV$ displays a large decrease. This is an intriguing finding, as $PR$ and $EV$ are typically correlated, yet they exhibit distinct behavior when stopwords are included in the analysis. Once again, the local strategy does not strongly affect this analysis. Regarding the variability ratio, apart from PR and EV, the values decrease significantly, indicating that the metrics become less dependent on syntax when stopwords are filtered. 

We now shift our analysis to metrics based solely on the most frequent words in the text. The results are presented in Table \ref{tab:comparision-strategy-part2}. 
For shorter text lengths (200 tokens), the informativeness improves significantly when stopwords are filtered for $L^*$, $CC^*$, $B^*$, and $EV^*$. However, informativeness slightly decreases for $C^*$ and drops considerably for $PR^*$. The variability ratio decreases in most cases when applying the global strategy, except for $PR^*$ and $EV^*$. 

\begin{table}[ht]
    \centering
    \caption{Comparison of informativeness and variability metrics among the Original network (O), Global strategy (G), and Local strategy (L), with and without stop-words filtering (SW). High values of informativeness measures in networks incorporating virtual edges are highlighted, along with their respective variability.}
    \label{tab:comparision-strategy-part2}
    %\small
    \begin{tabular}{lcc|cc|cc||cc|cc|cc}
        \hline
        &\multicolumn{6}{c}{Informativeness}&\multicolumn{6}{c}{Variability}\\
        \cline{2-13}
        &$\mathcal{O}$&$\mathcal{O}_F$&$\mathcal{G}$&$\mathcal{G}_F$&$\mathcal{L}$&$\mathcal{L}_F$&$\mathcal{O}$&$\mathcal{O}_F$&$\mathcal{G}$&$\mathcal{G}_F$&$\mathcal{L}$&$\mathcal{L}_F$\\
        \cline{2-13}
        &\multicolumn{12}{c}{Text size 200}\\
        \cline{2-13}
        L* & 61.0 & 86.0 & 61.0 & 95.0 & 65.0 & 97.0& 1.8& 0.9& 1.8& 0.8& 1.3& 0.8\\
        C* & 64.0 & 59.0 & 64.0 & 59.0 & 69.0 & 64.0& 1.7& 0.8& 1.7& 0.8& 1.4& 0.8\\
        CC* & 79.0 & 38.0 & 79.0 & 100.0 & 79.0 & 99.0& 1.7& 0.9& 1.7& 1.6& 1.7& 1.8\\
        B* & 39.0 & 61.0 & 59.0 & 100.0 & 59.0 & 100.0& 1.8& 1.2& 1.9& 1.3& 1.9& 1.3\\
        PR* & 73.0 & 61.0 & 88.0 & 61.0 & 88.0 & 78.0& 2.1& 1.5& 1.5& 1.5& 1.5& 1.1\\
        EV* & 64.0 & 98.0 & 80.0 & 98.0 & 80.0 & 97.0& 2.1& 1.4& 1.0& 1.4& 1.2& 1.3\\
        \hline
        &\multicolumn{12}{c}{Text size 400}\\
        \cline{2-13}
        L* & 68.0 & 98.0 & 68.0 & 100.0 & 70.0 & 100.0& 1.7& 0.9& 1.7& 0.7& 1.2& 0.6\\
        C* & 74.0 & 69.0 & 78.0 & 69.0 & 79.0 & 69.0& 1.7& 0.8& 1.3& 0.8& 1.2& 0.8\\
        CC* & 91.0 & 22.0 & 91.0 & 99.0 & 91.0 & 99.0& 2.1& 0.8& 2.1& 1.2& 2.1& 1.5\\
        B* & 45.0 & 87.0 & 64.0 & 100.0 & 64.0 & 100.0& 1.4& 0.8& 1.3& 0.9& 1.3& 0.9\\
        PR* & 91.0 & 63.0 & 99.0 & 63.0 & 99.0 & 83.0& 1.1& 0.9& 1.0& 0.9& 1.1& 0.9\\
        EV* & 77.0 & 96.0 & 93.0 & 96.0 & 93.0 & 95.0& 1.6& 1.0& 1.0& 1.0& 1.1& 1.0\\
        \hline
        &\multicolumn{12}{c}{Text size 800}\\
        \cline{2-13}
        L* & 75.0 & 99.0 & 81.0 & 100.0 & 85.0 & 100.0& 1.6& 0.8& 0.9& 0.5& 1.2& 0.6\\
        C* & 81.0 & 72.0 & 85.0 & 72.0 & 89.0 & 72.0& 1.7& 0.5& 1.4& 0.5& 1.3& 0.5\\
        CC* & 98.0 & 17.0 & 98.0 & 95.0 & 98.0 & 96.0& 2.0& 0.8& 2.0& 1.7& 2.0& 1.2\\
        B* & 49.0 & 98.0 & 86.0 & 99.0 & 86.0 & 99.0& 1.6& 0.6& 1.5& 0.6& 1.5& 0.6\\
        PR* & 100.0 & 54.0 & 100.0 & 57.0 & 100.0 & 75.0& 1.7& 0.8& 1.7& 0.7& 1.7& 0.6\\
        EV* & 80.0 & 93.0 & 95.0 & 93.0 & 97.0 & 94.0& 2.2& 0.9& 0.6& 0.9& 0.6& 0.9\\
        \hline
        &\multicolumn{12}{c}{Text size 1000}\\
        \cline{2-13}
        L* & 78.0 & 100.0 & 78.0 & 100.0 & 86.0 & 100.0& 1.7& 0.7& 1.7& 0.7& 1.2& 0.7\\
        C* & 80.0 & 65.0 & 83.0 & 76.0 & 88.0 & 76.0& 1.7& 0.6& 1.5& 0.5& 1.3& 0.5\\
        CC* & 97.0 & 8.0 & 97.0 & 96.0 & 97.0 & 95.0& 2.5& 1.0& 2.5& 1.3& 2.5& 1.2\\
        B* & 60.0 & 95.0 & 85.0 & 98.0 & 85.0 & 98.0& 1.4& 0.8& 1.3& 0.8& 1.3& 0.7\\
        PR* & 98.0 & 62.0 & 100.0 & 64.0 & 100.0 & 73.0& 2.0& 0.9& 1.6& 0.9& 1.9& 0.8\\
        EV* & 78.0 & 95.0 & 95.0 & 95.0 & 97.0 & 95.0& 2.3& 1.0& 0.8& 1.0& 0.9& 1.0\\
        \hline
    \end{tabular}
\end{table}

When analyzing larger text segments ($1,000$ tokens), excluding stopwords improves informativeness only for $L^*$ and $B^*$. Significantly, the informativeness of $PR^*$ decreases. Regarding the variability ratio, excluding stopwords generally leads to decreased values. The ability to capture linguistic features clearly shifts from syntax to semantics for $L^*$, $C^*$, and $B^*$. Interestingly, despite a significant decrease in the variability ratio for $CC^*$, this metric continues to capture syntactical features.

All in all, the results show that filtering stopwords may have a strong affect in both informativeness and variability ratio. This may happen independently of the chosen thresholding strategy. %\textcolor{red}{discuss here.}

\section{Conclusion}

In this paper, we analyze the statistical properties of enriched complex networks for text analysis. Although enriched co-occurrence networks have been applied in various contexts, the impact of including virtual edges has not yet been studied. Our focus was on two main properties: (i) \emph{informativeness}, which refers to a metric's ability to distinguish between meaningful and nonsensical texts; and (ii) \emph{variability ratio}, which reflects a metric's ability to capture syntactic or semantic variations. 

Our analysis revealed several interesting results. For instance, the addition of virtual edges can enhance the informativeness of certain metrics, such as the average shortest path—particularly in shorter texts—and closeness centrality. However, caution is warranted, as some metrics, like the clustering coefficient in short texts, may experience a decline in informativeness. Interestingly, other metrics, such as betweenness, appear to be unaffected by the inclusion of virtual edges in terms of informativeness.

Regarding the variability ratio analysis, we found that the inclusion of virtual edges in short texts increases the sensitivity of the average shortest path to semantics. Other metrics, such as eigenvector centrality, showed little effect from virtual edges and did not exhibit a clear dominance between semantic or syntactic features. Interestingly, the nature of certain metrics can change depending on the number of edges added. For instance, in the case of clustering and betweenness in short texts, the variability ratio decreases with the inclusion of virtual edges, shifting sensitivity from syntax to semantics. All the results confirm that the inclusion of virtual edges can play a significant role in determining which metric is best suited for use in specific NLP applications.

While our focus was on informativeness, this work could be extended to consider other types of noise in texts beyond shuffled words. For example, we could analyze the impact on metrics when authors attempt to conceal their identity in texts, such as during anonymization. This could involve changes in writing style or the introduction of errors, including synonym or antonym replacement, as well as other forms of word substitution. Additional possibilities include simulating typos, sentence fragmentation, or introducing ambiguity.  All of these strategies could prove useful in the task of author masking identification.

\section*{Acknowledgments}

This study was financed in part by the Coordenação de Aperfeiçoamento de Pessoal de Nível Superior - Brasil (CAPES) - Finance Code 001. Diego R. Amancio acknowledges financial support from CNPq (Grant no. 304026/2018-2, 311074/2021-9)  and FAPESP (Grant no. 2020/06271-0).

\bibliographystyle{ieeetr}
\bibliographystyle{abbrv}
%

%\bibliography{ref}

\newpage

\appendix
\section*{Supplementary Information}

Figures \ref{fig:infovar-local_strategy} and \ref{fig:infovar-local_strategyast} illustrate how the informativeness and variability ratios of the metrics behave when applying the local strategy for pruning semantic edges.

\renewcommand{\thefigure}{S\arabic{figure}}
\setcounter{figure}{0}

\begin{figure}[hbtp!]
    \centering
    \includegraphics[width=0.95\linewidth]{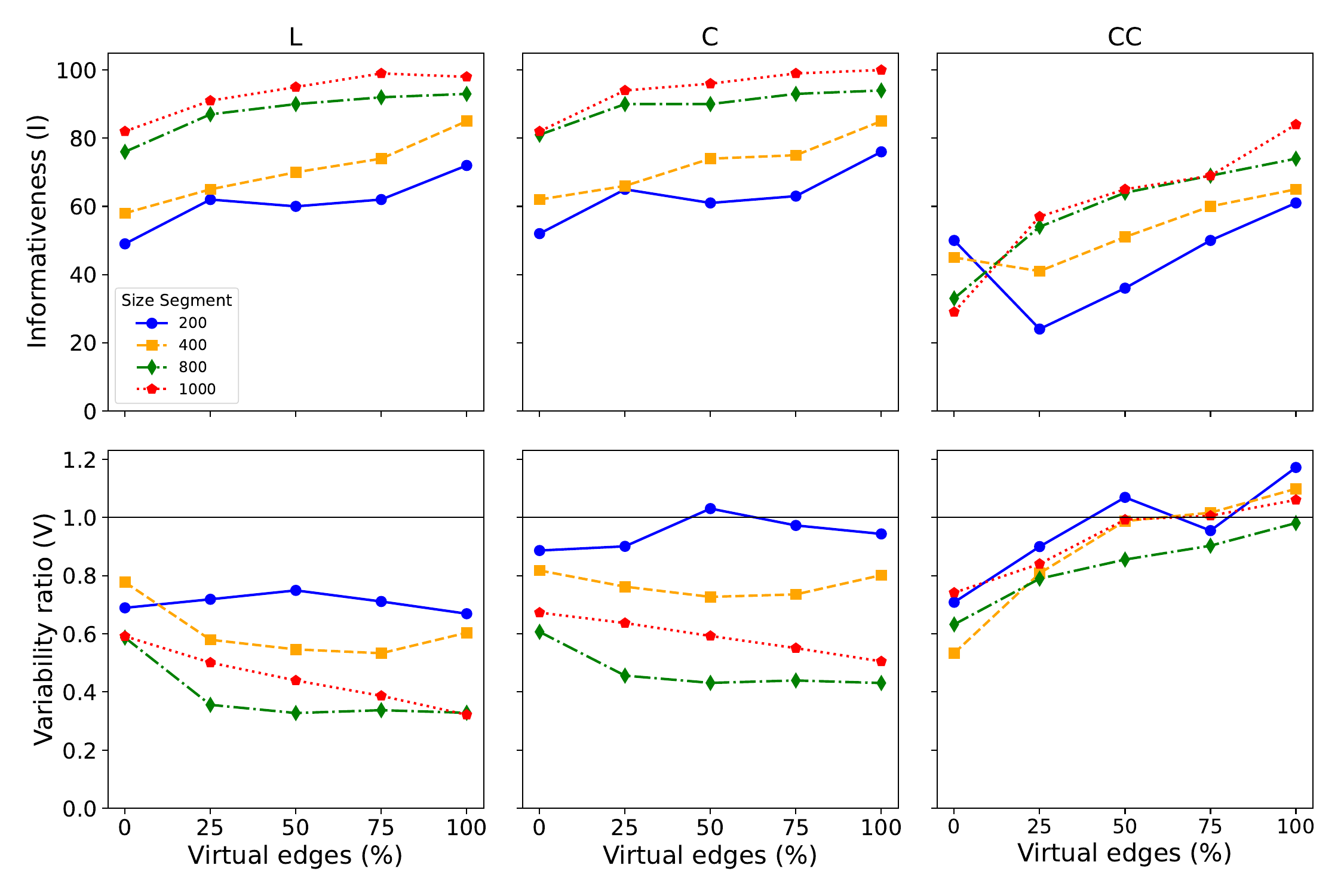}
    \includegraphics[width=0.95\linewidth]{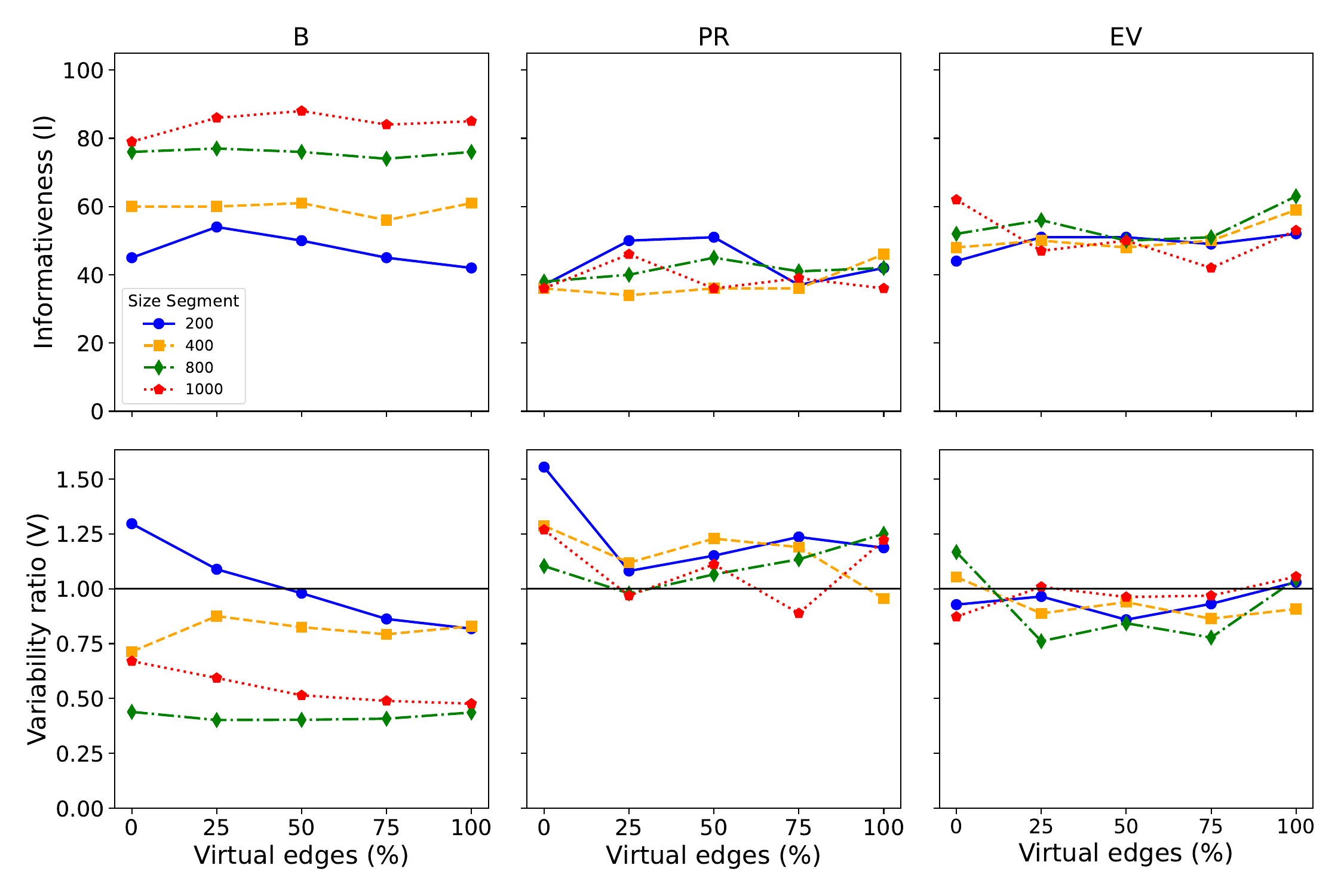}
    \caption{Local Strategy: Distribution of Informativeness and Variability measures for the Average Shortest Path (L), Closeness Centrality (C), Clustering Coefficient (CC), Betweenness Centrality (B), PageRank (PR), and Eigenvector Centrality (EV), with the addition of virtual edges in networks generated with variate text sizes and with filtering stop-word.}
    \label{fig:infovar-local_strategy}
\end{figure}

\label{app1}


\begin{figure}[hbtp!]
    \centering
    \includegraphics[width=0.95\linewidth]{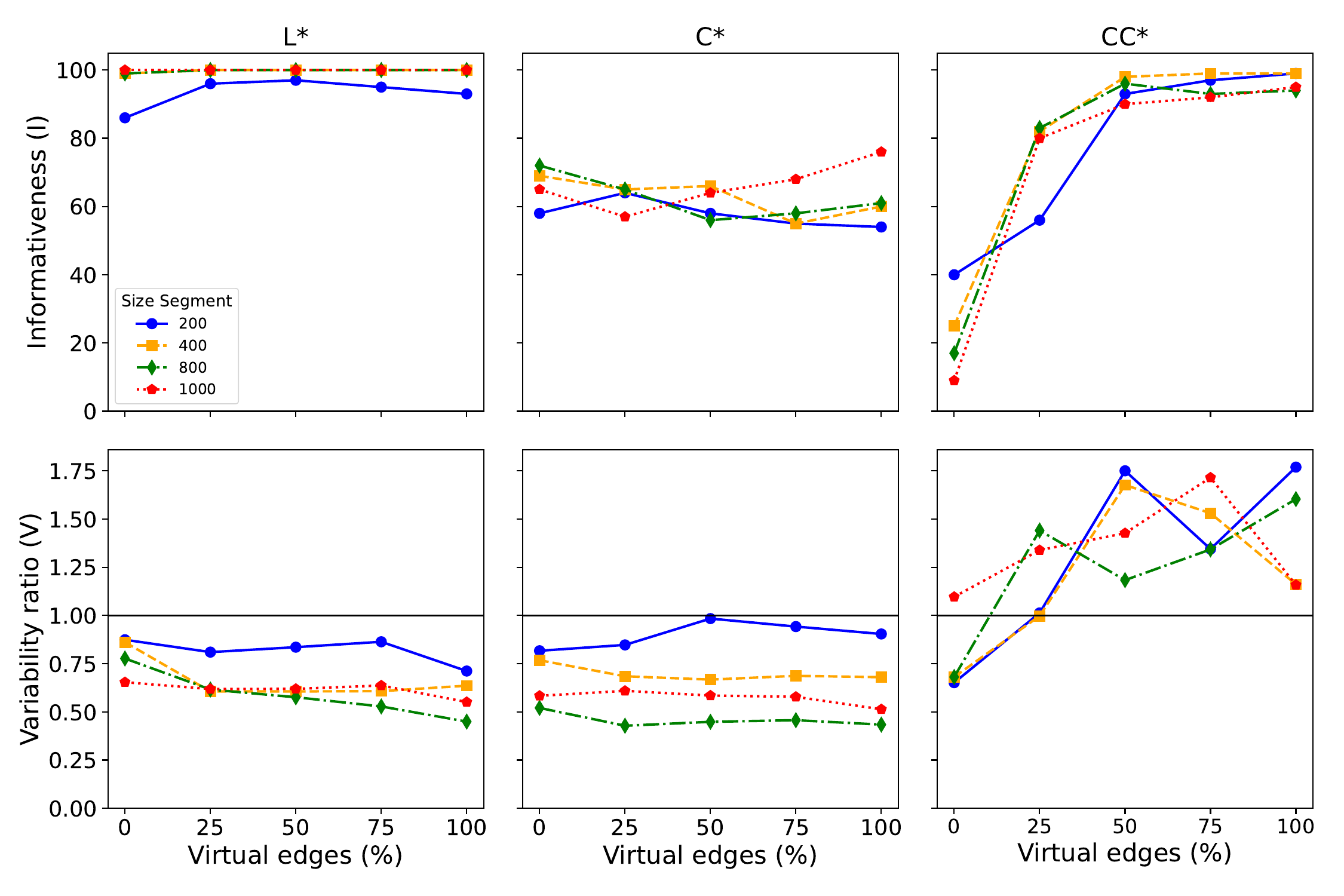}
    \includegraphics[width=0.95\linewidth]{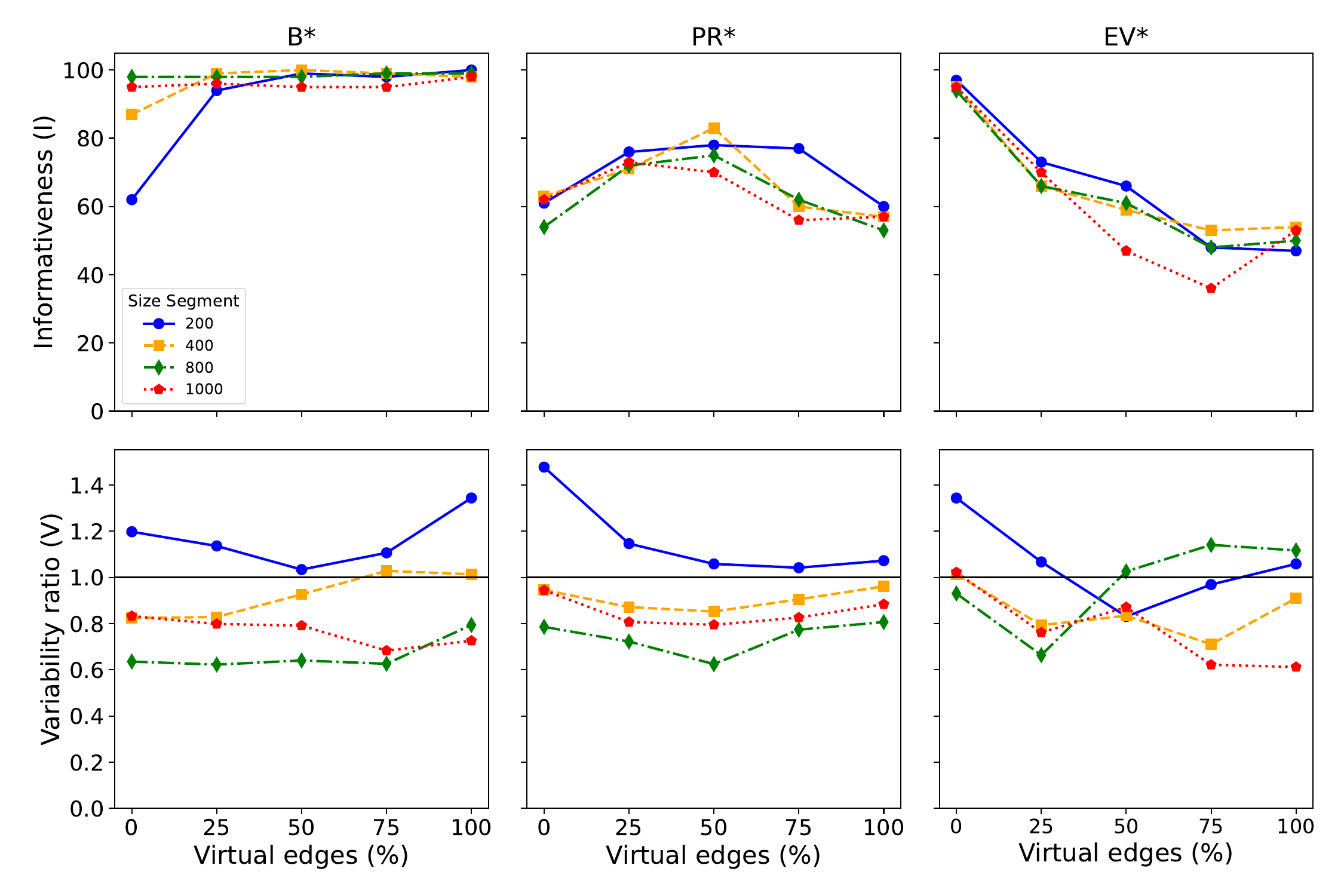}
    \caption{Local Strategy: Distribution of Informativeness and Variability measures for the Average Shortest Path (L*), Closeness Centrality (C*), Clustering Coefficient (CC*), Betweenness Centrality (B*), PageRank (PR*), and Eigenvector Centrality (EV*), with the addition of virtual edges in networks generated with variate text sizes and with filtering stop-word.}
    \label{fig:infovar-local_strategyast}
\end{figure}

\end{document}